\documentclass[10pt,twocolumn,letterpaper]{article}

\usepackage{cvpr}              

\usepackage{graphicx}
\usepackage{amsmath}
\usepackage{amssymb}
\usepackage{booktabs}
\usepackage{footmisc}
\usepackage{nicefrac}
\usepackage{color}
\usepackage{comment}
\usepackage{multirow}
\usepackage{footmisc}

\usepackage{comment}

\usepackage[pagebackref,breaklinks,colorlinks]{hyperref}

\makeatletter
\newcommand{\printfnsymbol}[1]{%
  \textsuperscript{\@fnsymbol{#1}}%
}
\makeatother

\usepackage[capitalize]{cleveref}
\crefname{section}{Sec.}{Secs.}
\Crefname{section}{Section}{Sections}
\Crefname{table}{Table}{Tables}
\crefname{table}{Tab.}{Tabs.}

\def\cvprPaperID{*****} 
\def\confName{CVPR}
\def\confYear{2022}

\begin{document}

\title{Augmenting Zero-Shot Detection Training with Image Labels}

\author{Katharina Kornmeier\thanks{equal contribution}\\
TU Darmstadt\\
\and
Ulla Scheler\printfnsymbol{1}\\
TU Darmstadt\\
\and
Pascal Herrmann\\
TU Darmstadt\\
}

\maketitle
\begin{abstract}
Zero-shot detection (ZSD), i.e., detection on classes not seen during training, is essential for real world detection use-cases, but remains a difficult task. Recent research attempts ZSD with detection models that output embeddings instead of direct class labels. To this aim, the output of the detection model must be aligned to a learned embedding space such as CLIP. However, this alignment is hindered by detection data sets which are expensive to produce compared to image classification annotations, and the resulting lack of category diversity in the training data.
We address this challenge by leveraging the CLIP embedding space in combination with image labels from ImageNet. Our results show that image labels are able to better align the detector output to the embedding space and thus have a high potential for ZSD. Compared to only training on detection data, we see a significant gain by adding image label data of 3.3 mAP for the 65/15 split on COCO on the unseen classes, i.e., we more than double the gain of related work.\\
This work was completed as an Integrated Project at TU Darmstadt.

\end{abstract}
\section{Introduction}
\label{sec:intro}
Object detectors for real world applications should be able to recognize a large amount of different classes. But in the conventional approach of object detection, every single class the detector should recognize has to be already represented in the training data set. This approach is quite limiting since not every one of these classes is known beforehand. This limitation becomes especially apparent due to the fact that commonly used object detection data sets are quite limited in the amount of different classes.\\
Zero-shot detection (ZSD) \cite{zeroshot} aims to overcome this limitation. The goal is to train the detector on a given set of classes (seen classes), but then be able to not only recognize the seen classes, but also classes that are not present during training (unseen classes). This way, zero-shot detectors are not limited to the classes the model was trained on.\\
One way to do ZSD is to use an embedding space to get a correspondence between images and labels. To make a classification prediction, a model using this approach predicts an embedding vector instead of directly predicting class probabilities. The correspondence through the embedding space can then be used to relate the embedding vector to the class prediction. This way, these models are no longer restricted by the number of classes, only by the expressiveness of the embedding space and how well the model is aligned to it. Recent ZSD models are often trained on either the 48/17 COCO split \cite{zeroshot} or the 65/15 COCO split \cite{65split}, meaning that only 48 and 65 classes are used for training, respectively. Therefore we believe that the alignment to the embedding space can be improved by presenting a larger variety of object categories.\\
One way how conventional object detection can be improved is to incorporate image classification data into the object detection training pipeline \cite{yolo9000}.
This way, the detector is able to detect a larger amount of object categories. But since this approach is proposed for conventional object detectors, class probabilities are predicted directly and the model is therefore still limited to the classes it was trained on.\\
In this paper, we use the idea to augment the training of an object detector with image classification data and apply it to the task of ZSD. Similar to previous work, we change the head of a single-stage detector, e.g., YOLOX \cite{yolox}, to predict embedding vectors instead of class probabilities. We then align our model to the CLIP \cite{clip-paper} embedding space by training on the 65/15 COCO split \cite{65split}. Moreover, we augment the training with ImageNet1000 \cite{1000ImageNet} classification data to improve the alignment of our model to the CLIP embedding space. We show that this improves the performance of our model significantly.  Furthermore, we propose an unseen/seen split for ImageNet1000, since using images from ImageNet that are part of unseen COCO classes would violate the ZSD setup.\\

\section{Related Work}

\label{sec:related_work}
\subsection{Object detection}
\label{sec:object_detection}
Object detection is the task of finding the positions of objects in an image by assigning an axis-aligned bounding box, and classifying them \cite{surveyObjectDetection}. It differs from image classification mainly in two ways: 1) More than one target object can be present in an image and 2) the locations of the objects have to be determined, not just the correct label. This makes object detection a more difficult task than image classification.\\
One approach to build an object detection model is to use a convolutional neural network (CNN) architecture. Even though current research moves in the direction of using Transformer-based \cite{transformers} models \cite{DETR,DeformableDETR}, we focus on CNN-based models.\\
The backbone can be the feature extractor of any desired image classification model, which is usually also pretrained on an image classification task. Here, the feature extractor is the image classification model without the classification layer.\\
Currently, there are two types of architectures: single-stage detectors and two stage detectors. Two-stage detectors \cite{RCNN,fastRCNN,fasterRCNN,SPP-Net, FPN, R-FCN, DetectoRS} perform object detection in two different steps. In the first step, region proposals are generated based on the feature map from the backbone. They consist of object proposals, i.e., assumed positions of objects, as well as an objectness score, which is a measure between 0 and 1 that determines whether the object proposal contains an object (values close to 1) or background (values close to 0). The region proposals are then used for classification and refinement of the localization in the second step. Due to this separate approach, two-stage detectors reach the highest accuracy rates but are usually slower. \cite{twostagevsonstage}.\\
Single-stage detectors \cite{yolox, SSD, yolo, RetinaNet, EfficientDet, SwinTransformer} only use a single step to classify and localize objects. The features from the backbone are fed into an object detection head. Here, bounding boxes, classification labels and objectness scores are calculated. The head can be decoupled, meaning that the network branches into independent layers, e.g., one for bounding boxes and objectness and one for the classification labels. Single-stage detectors are much faster but only reach lower accuracy rates. \cite{twostagevsonstage}.\\
Object detectors can also be distinguished by using anchors or being anchor-free. Anchors can be seen as priors on bounding box sizes and shapes. The number of anchors is predefined and they are precomputed before the training of the detector. Anchor-based methods, as proposed in \cite{fasterRCNN}, do not propose bounding boxes directly. Instead, for every anchor, they predict a deviation from the corresponding anchor. Anchor-free methods do not use anchors and predict bounding box positions directly.\\
One recent line of work on single-stage detectors is the YOLO-series \cite{yolo,yolo9000,yolov3,yolov4}, which puts great emphasis on real-time capabilities. Redmon et al. \cite{yolo} propose YOLO, which tackles object detection as a regression rather than a classification problem. They divide the image into a grid and predict a number of bounding boxes as well as class probabilities for every grid position. These are then combined to make the final prediction. Additionally, YOLO is pretrained on ImageNet1000 \cite{1000ImageNet}.\\
Redmon amd Farhadi \cite{yolo9000} identify problems with YOLO, namely bad localization as well as low recall compared to two-stage detectors. They propose several changes to YOLO, while keeping its runtime in mind, introducing YOLOv2. They include batch normalization \cite{batchNormalization} to improve convergence. Furthermore, they pretrain their model on ImageNet1000 \cite{1000ImageNet} on the full resolution of the images, while YOLO only uses half the image size. Moreover, YOLOv2 uses anchors, to make predictions while YOLO is an anchor-free method. This improves the recall of YOLOv2. Finally, they propose Darknet-19 as their backbone, while YOLO uses a custom CNN. On top of YOLOv2, they also propose YOLO9000, to increase the expressiveness of their method to more than 9000 classes. To achieve this, they train jointly on COCO \cite{coco} and ImageNet \cite{1000ImageNet}. When the detector sees an image from COCO (bounding box annotations + classification labels), the complete loss is used to adjust the model weights. When the detector sees an image from ImageNet (only image labels), only the classification loss is used.\\
Next, Redmon and Farhadi propose YOLOv3 \cite{yolov3}, which only introduces some minor changes to YOLOv2, e.g. changing the backbone to Darknet-53 and predicting bounding boxes at multiple scales. These changes especially improve the average precision for small objects, something the previous YOLO versions struggle with.\\
Ge et al. propose YOLOX \cite{yolox}, which builds upon YOLOv3. They move back to the roots of YOLO by making YOLOX anchor-free again. Furthermore, they use a decoupled head, i.e., classification predictions and bounding box predictions are done by separate branches of the network architecture. Additionally, they use strong data augmentation like MixUp \cite{mixup} and Mosaic\footnote{\url{https://github.com/ultralytics/yolov3}} which allows them to train YOLOX from scratch without ImageNet \cite{1000ImageNet} pretraining. Finally, they introduce the label assignment strategy SimOTA, based on OTA \cite{OTA}.\\
Ramanathan et al. \cite{dlwl} propose an approach to incorporate weakly supervised data, i.e., images with a set of labels but no bounding boxes, into the training of an object detection model. They assign bounding boxes to the weakly supervised images based on a constraint optimization problem. This improves the average precision especially for classes that have a low number of bounding box annotations.\\
In this work, we build upon YOLOX and incorporate the idea to augment the training with image classification data to increase the expressiveness of our approach.\\

\subsection{Zero-shot detection}
\label{sec:zero_shot_learning}
Zero-shot detection (ZSD) \cite{zeroshot} is a special form of object detection. Contrary to object detection, not every object class a ZSD detector should categorize during testing is present in the training data set. Those classes are called unseen classes while the classes present during training are called seen classes.\\
To evaluate the performance of a ZSD model, a data set is needed which contains only seen classes in the training data set and unseen classes in the test data set. For this purpose, commonly used object detection data sets are split in seen and unseen classes. One example is the COCO \cite{coco} data set. Currently, two splits are widely used: The 48/17 split \cite{zeroshot} and the 65/15 split \cite{65split}.\\
For the 48/17 split, word vector embeddings are calculated for all classes of COCO. They are then clustered into 10 clusters using cosine similarity as a measure. From each cluster, 80\% of the classes are randomly assigned to the seen classes, while the other 20\% are assigned to the unseen classes. But only those classes are used, which have a synset associated with them in the WordNet hierarchy \cite{wordnet}, resulting in the 48 seen classes and 17 unseen classes.\\
Rahman et al. \cite{65split} identify some problems with the 48/17 split. On the one hand, the 48/17 split only contains 65 classes while COCO contains 80 classes. Therefore, they do not utilize the full potential of COCO. On the other hand, the chosen clustering does not guarantee a diverse set of unseen classes. To tackle these issues, they proposed the 65/15 split. Based on the total number of classes, the classes of each superclass from COCO are sorted in ascending order. Then, 20\% of the rare classes from each superclass are selected as unseen classes, the rest as seen classes. This ensures a diverse nature of unseen classes while utilizing the complete COCO data set. Because of these advantages, we only use the 65/15 split in this paper.\\
The problem of ZSD is especially interesting due to the fact that object detection data sets are more difficult to annotate than image classification data sets, because not only a classification label has to be provided but also the location of the object. Therefore, object detection data sets are quite small compared to object image classification data sets. The 2014 release of COCO \cite{coco} contains roughly 83,000 images divided into 80 classes. Compared to this, the ImageNet1000 \cite{1000ImageNet} data set contains roughly 1,200,000 images divided into 1000 classes.\\
One way to tackle ZSD is to use an embedding space \cite{zeroshot}. The idea is to not directly predict a classification for a given image but rather an embedding vector, which is an abstract representation of the classification label. A currently used embedding space is the CLIP embedding space \cite{clip-paper}. More embedding spaces, like ALIGN \cite{ALIGN}, GloVE \cite{gloVE}, ConSE \cite{conSE}, and DeViSE \cite{devise}, exist but we focus on CLIP in this paper.\\
CLIP consists of an image encoder and a text encoder which both output n-dimensional embedding vectors. During training, a contrastive loss \cite{contrastiveLoss} is used, meaning that image embeddings are computed using the image encoder and the corresponding text labels are fed through the text encoder to compute text embeddings. Then, both the image encoder and the text encoder are tuned in a way to maximize the cosine similarity of corresponding image text pairs while minimizing the cosine similarity of non-corresponding image text pairs. This way, images are related to their labels through the embedding space.\\
CLIP was developed for the task of zero-shot classification, i.e., predicting a label for an image of unseen classes. The idea is, that the trained embedding space is not only able to relate images and labels of seen classes but also of unseen classes. To make a zero-shot classification prediction, the labels of unseen classes are fed through the text encoder, generating embedding vectors. Then, an image of an unseen class is fed through the image encoder, generating an embedding vector as well. To make a prediction, the cosine similarity between the image embedding vector and all the text embedding vectors of seen and unseen classes is calculated. The label embedding with the highest cosine similarity is then the prediction, which ideally corresponds to the correct label of the unseen class. This way, CLIP is not restricted to the classes it was trained on, enabling zero-shot classification.\\
This idea can be used for ZSD as well.\\
Xie and Zheng propose ZSD-YOLO \cite{zsd-yolo}, which builds upon YOLOv5\footnote{\url{https://github.com/ultralytics/yolov5}}, a single-stage detector. They alter the detection head in a way to output an embedding vector instead of class probabilities in addition to bounding box and objectness outputs. During training, they align this embedding vector to the output of the pretrained CLIP \cite{clip-paper} text encoder as well as to the pretrained image encoder. This allows them to use CLIP as a classifier and make predictions for unseen classes. Additionally, they propose a self-labeling strategy to increase the number of examples the model can learn from. For this, they use their model, only trained on seen classes, and run inference on the images of the training set. Non-maximum suppression is applied, where ground truth bounding boxes are prioritized. The new bounding boxes are then treated as ground truth boxes in further training. Before they train their model on the ZSD task, they train the unaltered YOLOv5 model only on the seen classes.\\
Another approach is ViLD \cite{vild-paper}, proposed by Gu et al., which builds upon maskRCNN \cite{maskRCNN}, a two-stage detector. They work on the task of zero-shot instance segmentation, where a pixel-wise mask for unseen classes is predicted, instead of just bounding boxes. But since implicitly bounding boxes are predicted as well, we mention this method here. They alter the classification part of their model to an embedding vector and align it to the text embeddings and image embeddings from CLIP \cite{clip-paper}. For the text embedding part of their model, they introduce a background category for predictions that do not match any of the ground truth labels. The background embedding is learned during training. The text embedding part is trained only using class names of seen classes, while the image part is also trained on embeddings of unseen classes. Additionally, they introduce an ensembling approach, where they weight the predictions of the text part and the image part based on if a class is seen or unseen. ViLD is trained from scratch without any pretraining.\\
Zareian et al. \cite{ovr-cnn} propose OVR-CNN. They pretrain their model on an image-caption data set to learn an embedding space. They then finetune their model on an object detection data set which only contains some of the classes of the image-caption data set. By this, they are able to also make bounding box predictions for the classes from the image-caption data set, achieving zero-shot predictions.\\ 
Our approach follows this line of work. We build upon YOLOX and change its classification output to an embedding vector. But in contrast to ViLD and ZSD-YOLO, we do not align to the image embeddings from CLIP but only to the text embeddings. Additionally, we follow the idea of YOLO9000 and augment our training with ImageNet1000 classification data to get a better alignment to CLIP.
\section{Method}
\label{sec:method}
\subsection{Overview}

A classic detection model directly outputs probabilities for class labels and then compares them to a one-hot encoded target vector. What makes it possible for our model to detect unseen classes, is that our model outputs an embedding of a class label. The idea is that learning the embeddings for the seen classes during the training implicitly teaches the model the embeddings for an unseen class.\\
We align YOLOX, a single-stage detector, to the CLIP embedding space by training on a detection data set $\mathcal{D}_D$, e.g., COCO, and a classification data set $\mathcal{D}_C$, e.g., ImageNet. This requires three steps: 1) changing the output of the classification head to output embeddings,  2) changing the classification loss based on similarity to the CLIP embedding and 3) a training strategy to include the classification data in the detection training pipeline.

\subsection{Changes to YOLOX}
The YOLOX detection model is a single-stage detector with a favorable accuracy-speed trade-off and a decoupled head, i.e., one head for objectness and bounding boxes and one head for classification predictions \cite{yolox}. We thus only change the classification head while the objectness and bounding box head remains unchanged. In its original version, the classification head of YOLOX outputs logits – one probability for each possible class. We change the output to be of the same dimension as the embedding space we align to.

\subsection{Loss with text embeddings}\label{embeddings-loss}
Our goal is to align the output of the detection model to the CLIP text embeddings. In other words: We input an image of class $i \in \{1, …, n\}$ into the detection model, and we input the name of the class $i$ into a text prompt to the CLIP text encoder. We receive a predicted embedding $e$ and a target embedding $c_i$, respectively, and we want them to be as close as possible.
Before we start the training, we once calculate the CLIP text embeddings for every of the $n$ classes we use during training. To make these single-word labels more similar to the sentences CLIP was trained on, the class names are enveloped by a text prompt, before they are encoded. We use several different text prompts and average them to receive a target embedding. We concatenate these target embeddings to a matrix which functions as a fixed linear classifier $C$ that remains unchanged by training/testing and is used every time the model makes a prediction.\\
Given an image, the model predicts bounding boxes and objectness as usual as well as a classification output in form of an embedding. For each predicted bounding box, we calculate the cosine similarities between the predicted embedding $e$ and every embedding $c_i$ in $C$. We pass the result through a softmax with temperature scaling with temperature $\tau$. The temperature scaling is necessary as the cosine similarity between the embeddings is non-zero. Thus without temperature scaling, the correct embedding would result in a loss value incorrectly significantly different from 0. The temperature scaling allows for a relaxation that makes it possible to favour even small differences in similarity.
In summary, the similarity vector $s$ is computed as follows:

$$
\mathbf{s}=\operatorname{softmax}\left(\frac{e^{\top} \mathcal{C}}{\|e\|\|\mathcal{C}\|} \exp(\tau)\right)
$$\label{softmax-loss}

The vector $s$ is thus of size $n$ and contains the probability of the respective bounding box showing each of the $n$ possible classes. This is similar to the logits output by classic detection models.
In a last step, we use a cross entropy loss to compare the vector $\mathbf{s}$ 
with the target label $t \in \{1, …, n\}$:

$$
\mathcal{L}=\mathcal{L}_{\mathrm{CE}}(\mathbf{s}, t )
$$

\begin{figure*}
  \centering
  \begin{subfigure}{\linewidth}
      \includegraphics[width=\linewidth]{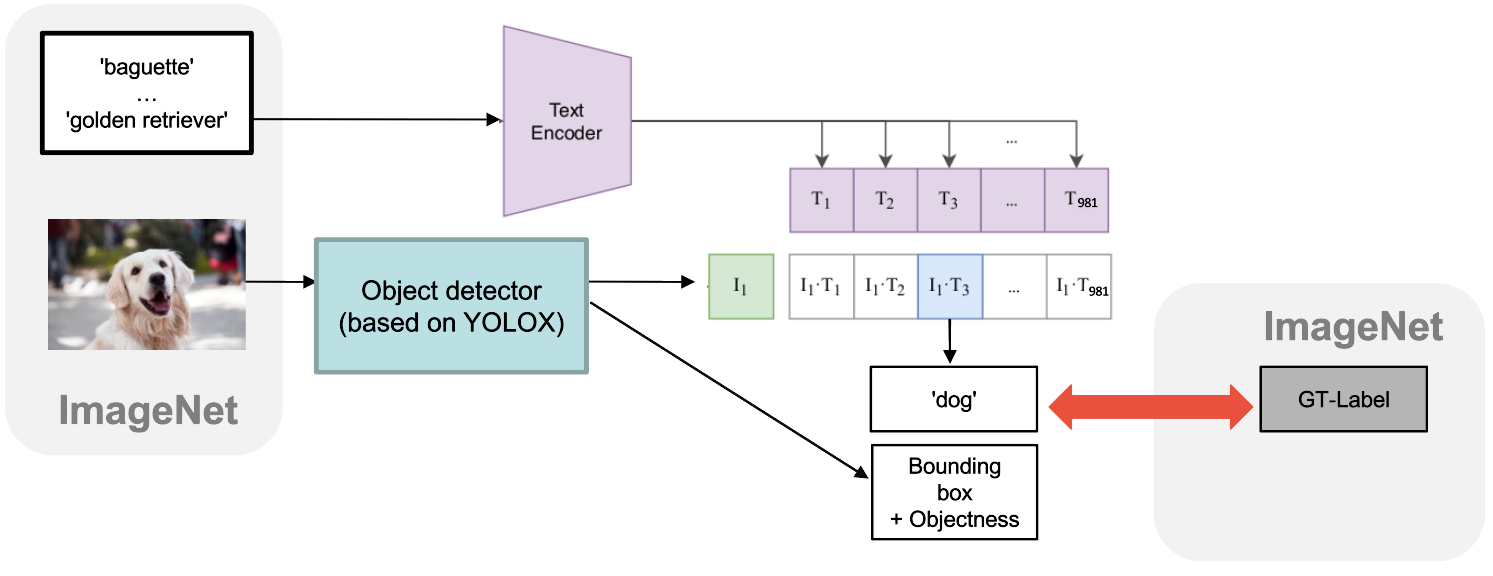}
  \end{subfigure}
  \caption{Loss calculation for ImageNet data. With an ImageNet image only the loss on the classification score is returned. Bounding box and objectness losses are set to zero. Note that we first train on COCO and then concurrently on COCO and ImageNet. Illustration adapted from \cite{clip-paper}.}
  \label{fig:loss-calculation}
\end{figure*}

\subsection{Joint training on detection and classification data}
When using the classification data set $\mathcal{D}_C$ we do not have ground truth bounding boxes and thus only backpropagate the classification loss. In contrast to \cite{yolo9000} we do not backpropagate an objectness loss on the classification data. An overview of this process can be found in figure \ref{fig:loss-calculation}.\\
Lacking ground truth bounding boxes in the classification data, we have to specify which of the many bounding box propositions of varying size and location to backpropagate the loss on. We adapt the procedure introduced by \cite{yolo9000} and proceed in the following two steps: 1) \textit{We filter all bounding box predictions with an objectness score smaller than a certain threshold $th_{obj}$.} One can think of this value as a sort-out criterion: We do not expect the objectness predictions to be as accurate for the classes of $\mathcal{D}_C$ as for the classes of $\mathcal{D}_D$ the model was trained on, but we expect them to be good enough at differentiating between general objects and the background. 2) \textit{In the case of $\mathcal{D}_C$ we choose the bounding box with the highest classification score for the ground truth class as ground truth bounding box.} In section \ref{embeddings-loss} we showed how we calculated the classification prediction by comparing each output embedding of the model to the matrix $\mathcal{C}$. The comparison is the same for $\mathcal{D}_D$ and $\mathcal{D}_C$, but with separate matrices $\mathcal{C}_{D}$ and $\mathcal{C}_{C}$.\\
Thus, implicitly, we make the assumptions that our alignment to the embedding space is already good enough, such that the highest classification score for the depicted object is received by a bounding box that indeed shows such an object. This assumption leads to several possible failure cases: 1) If there are multiple objects of the image label category in the image, we only backpropagate the loss on one bounding box. This case does not pose a problem, as we do not penalize the other bounding box predictions. 2) If there are multiple objects of different kinds, we might falsely backpropagate on a bounding box that shows a different object. In this case, the model would indeed learn from wrong information. We meet this case with pretraining on the detection data set, such that the predictions of the model are already as reliable as possible before we use the classification data. The process is further aided by the object centricity of many classification data sets where the ground truth object is featured prominently in the middle of the image with few - if any - other objects in the frame.

\section{Experiments}
\label{sec:experiments}
\subsection{Data and metrics}
We use CLIP \cite{clip-paper} as our embedding space, COCO \cite{coco} as our detection data set and ImageNet1000 \cite{1000ImageNet} as our classification data set.
We test the zero-shot detection abilities of our model by using the  65/15 COCO split of \cite{65split}, wherein the data is split into 65 classes seen during training (“seen classes”) and 15 classes we keep back for testing (“unseen classes”). Accordingly we also filter the 1000 classes of ImageNet \cite{1000ImageNet} by excluding all categories that are a part of, identical to, or a specific version of an unseen COCO class. We arrive at 981 seen ImageNet classes that we use during training (the exact classes we exclude are described in section \ref{appendix} of the appendix).\\
Following \cite{yolo9000}, we oversample COCO with a ratio of about 4:1 of ImageNet vs. COCO.
Adopting the convention of other zero-shot detection works we report both mAP and AR@100 at an IoU of 0.5, where AR@100 is the average recall given 100 detections per image.

\subsection{Implementation details}
As a base model we use YOLOX-l as implemented in the publicly available MMDetection repository\footnote{\url{https://github.com/open-mmlab/mmdetection}\label{mmdet}}.
The COCO images are of size $640 \times 640$~px. ImageNet images are choosen to be smaller than COCO images because objects in ImageNet images are usually larger within the image. ImageNet images are of size $320 \times 320$~px.
Unlike \cite{yolox} we do not use Mosaic and Mixup data augmentation during training due to time constraints.
We train with stochastic gradient descent on three NVIDIA GeForce RTX 2080 Ti GPUs with a batch size of 2 images per GPU. We accumulate the gradients over 4 batches to imitate a larger batch size and use a synced batch norm. Weight decay is set to 5e-4. We use SGD with a momentum is 0.9. The linear warm-up lasts for 5 epochs.
The learning rate follows a cosine schedule with an adapted initial learning rate of about 0.001. Samples of COCO and ImageNet can occur in the same batch. Since objectness and bounding box loss is only calculated based on the COCO images we scale the learning rate for objectness and bounding box loss according to the number of COCO images present in the batch.\\
As in the original CLIP paper\cite{clip-paper} the temperature is a learnable parameter, initialized to $\nicefrac{1}{0.07}$ and clipped with a maximum value of 100.\\
The target text embeddings are generated with the text encoder corresponding to the ViT-B/32 CLIP model\footnote{\url{https://github.com/openai/CLIP}\label{clip-git}}.  In its original version, the classification head of YOLOX outputs an 80-dimensional vector – one probability for each COCO class. We change the output to be of dimension 512 – the embedding dimension of the CLIP embedding space with CLIP ViT-B/32. To generate the text embeddings of the available classes, we insert the COCO or ImageNet label into seven sentence templates that proved to be most helpful for generating embeddings\footnote{\url{https://github.com/openai/CLIP/blob/main/notebooks/Prompt_Engineering_for_ImageNet.ipynb}\label{clip-embeddings-git}}, for example “a bad photo of the \{class\}.” The templates can be found in supplemental material \ref{appendix}. For COCO, we used the class names as they were. For ImageNet, we used the labels from a related work\footref{clip-embeddings-git}, where only one label is provided per category and labels were extended in case of homonyms. For example, 'kite' was specified as 'kite (bird of prey)'. Furthermore, we added '(dog)' to all dog classes.\\
 First, we pretrain our model on COCO to achieve reliable bounding boxes and objectness scores. Second, we train jointly on COCO and ImageNet. During the loss calculation for ImageNet data, we filter all bounding box predictions with an objectness score of less than 0.001.\\
 
During inference we use classification scores as confidence scores. YOLOX in contrast generates confidence scores by multiplying the classification scores with objectness scores. But just like ZSD-YOLO \cite{zsd-yolo} we observed that it is better to not factor in objectness scores when predicting unseen classes. To still filter out background, we apply an objectness threshold of 0.1. Further we apply a classification score cutoff of 0.01 and perform non-maximum supression with an IoU threshold of 0.5.

\subsection{Results}\label{sec:results}

\begin{table}
  \centering
  \begin{tabular}{lclc}
    \toprule
    Method & mAP & & AR@100\\
    \midrule
    ZSD-YOLO-text \cite{zsd-yolo} & 16.8 & & (66.0)\\
    ViLD-text\footref{video} & 14.71 & & 47.69 \\
    Ours (w/o image labels) & 14.7 & & 44.0\\
    \midrule
    ZSD-YOLO \cite{zsd-yolo} & 18.3 & \textbf{+1.5} & (69.5)\\
    Ours & 18.0 & \textbf{+3.3} & 45.8\\
    \bottomrule
  \end{tabular}
  \caption{mAP and AR (in \%) of different models on the unseen classes of the 65/15 COCO split. The recall for ZSD-YOLO-text is not representative because they used different threshold when evaluating recall compared to evaluating mAP. Aligning to the text embeddings of COCO and ImageNet is superior to aligning to text and image embeddings on COCO as the overall gain in mAP is higher. Lower baseline results of our model might stem from less parameter tuning during training and a smaller backbone model. Results for ViLD on the 65/15 COCO split with a ViT-B/32 CLIP model were not available.}
  \label{tab:main_results}
\end{table}

\begin{table}
  \centering
  \begin{tabular}{lcc}
    \toprule
    Method & mAP & AR@100\\
    \midrule
    Ours & 14.7 & 44.0\\
    Ours (longer training) & 14.1 & 43.0\\
    \bottomrule
  \end{tabular}
  \caption{mAP and AR (in \%) of our model on the unseen classes of the 65/15 COCO split. Both models are trained on COCO only. The first row is trained for 170 epochs and the second row for 220 epochs, showing that the gain from adding ImageNet is not from longer training on COCO.}
  \label{tab:no_gain_only_coco}
\end{table}

 \begin{table*}[t]
  \centering
  \begin{tabular}{lccccc}
    \toprule
    \multirow{2}{*}{Model} &
      \multicolumn{4}{c}{mAP across scales} &
      \multirow{2}{*}{AR@100}\\
      & s & m & l & all & \\
    \midrule
    YOLOX-l & 68.7 & 90.0 & 91.7 & 85.6 & 90.6\\
    Ours (w/o data augm.) & 35.1 & 57.5 & 61.6 & 52.7 & 66.3\\
    \bottomrule
  \end{tabular}
  \caption{AR and mAP of small (s), medium (m) and large (l) size objects on the seen classes of the 65/15 COCO split. Both models are trained on COCO only. The large gap in performance between our model and the unmodified YOLOX-l models is especially salient for small objects. The unmodified YOLOX-l model uses Mosaic and Mixup data augmentations during training, our model does not.
  }
  \label{tab:yolox_natur}
\end{table*}

\begin{figure*}[t]
  \centering
    \includegraphics[width=\linewidth]{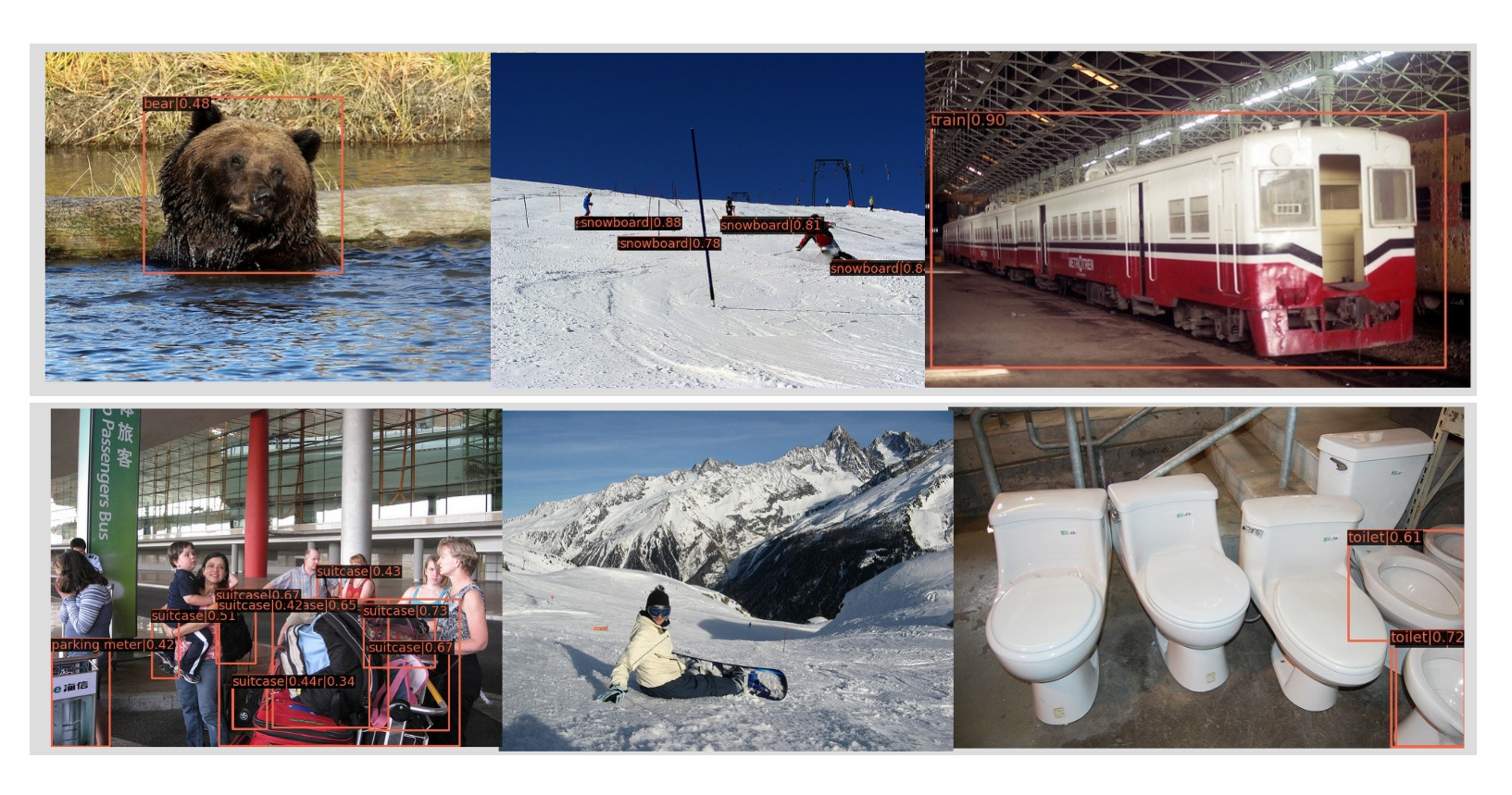}
  \caption{Qualitative results on unseen COCO classes. First row: Successful zero-shot detection on objects of varying size. Second row: Unsuccessful examples.}
  \label{fig:example_images}
\end{figure*}

Table \ref{tab:main_results} presents the results of our method in comparison to previous work.\\
We first evaluate mAP and recall of our model when trained only on COCO since we are interested in the gain by adding ImageNet data. We compare our COCO-only baseline with \mbox{ZSD-YOLO-text} \cite{zsd-yolo} and ViLD-text\footnote{\label{video}\url{https://www.youtube.com/watch?v=aA0r1M_NWhs}}. Both models also align to COCO and the text embeddings of the CLIP embedding space and are thus comparable with our baseline trained only on COCO.
Our COCO-only baseline is up to par with ViLD-text, but worse than ZSD-YOLO-text in mAP. The authors of ZSD-YOLO-text used different confidence thresholds when evaluating recall compared to evaluating mAP, misleadingly improving their recall. This prevents us from comparing to the recall of ZSD-YOLO-text. \\
Next, we contrast the results of our final model - which is aligned to the text embeddings of both COCO and \mbox{ImageNet1000} - to the full ZSD-YOLO model which is trained on COCO and aligns to the text as well as the image embeddings of the COCO data. Despite the fact that ZSD-YOLO reaches a higher mAP by 0.3 percentage points, the overall gain of 3.3 percentage points in mAP from adding ImageNet in our model is considerably higher than the gain of 1.5 percentage points in mAP from adding image embedding alignment in ZSD-YOLO. Table \ref{tab:no_gain_only_coco} shows that our improvements from adding ImageNet are not from the longer training iterations on COCO. Further training on the detection data set does not improve ZSD, instead finetuning on COCO seen classes leads to a decline on the unseen classes. This shows that our gain of 3.3 percentage points in mAP exclusively arises from augmenting the training with image labels. Thus, our improvement of adding image labels outperforms the improvement of adding image embeddings.

There are several factors that explain why our model is not able to outperform ZSD-YOLO in overall mAP, the most obvious being that we use only the second-to-largest model of the YOLOX family and thereby using fewer model parameters than ZSD-YOLO.
Another explanation lies in the data augmentations that we do not use. Both the reduction of the model size and the reduction of data augmentation is due to time and hardware constraints. To estimate the effect of data augmentation, we compare the results of an unmodified YOLOX-l model trained on COCO with our modified YOLOX model of the same size on the seen classes in table \ref{tab:yolox_natur}. For comparison to the unmodified YOLOX-l we downloaded the weights from a public repository\footnote{\url{https://github.com/open-mmlab/mmdetection/blob/master/configs/yolox/README.md}}.  mAP is considerably worse for the results of our model over all object sizes, but the case is especially severe for small objects. These models differ only in 1) changes in the classification head to make our model compatible with embeddings, as discussed in section \ref{sec:experiments}, and 2) the aforementioned left out data augmentations.
With an additional experiment that is presented in section \ref{sec:additional_experiments} of the appendix, we show that the changes in the classification head only have a small influence on the mAP. Thus the lower performance in contrast to the unmodified YOLOX-l model can at least in part be attributed to the missing data augmentations. Having more computational resources available, it is therefore useful to train with Mosaic and Mixup, such as the unmodified YOLOX-l does. Further training improvements could be achieved with the tuning of the learning rate and experimentation with larger batch sizes.

We depict some exemplary zero-shot detection results of our model in figure \ref{fig:example_images}. Failure cases include unknown views of objects that are correctly classified otherwise (snowboard), only recognizing a certain configuration of an object (toilets) as well as finding the right objects but employing too many bounding boxes (suitcases). 

\begin{figure}[t]
  \centering
    \includegraphics[width=\linewidth]{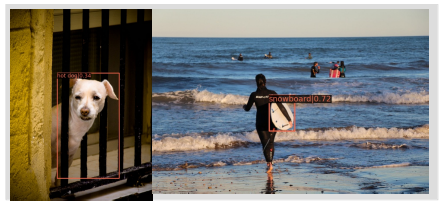}
      \caption{Detection results for unseen labels on images with seen classes. While a surfboard is visually similar to a snowboard (right image), a dog and a hot dog are similar because of their label.}
  \label{fig:example_images_better_labels}
\end{figure}

While visually similar objects can easily be confused by the detector, figure \ref{fig:example_images_better_labels} illustrates a new failure mode that is due to the underlying structure of label embeddings, like in the image where a dog is misclassified as a "hot dog". Obviously, detection as well as classification data sets were labeled with the mere goal of being understood by humans, and are thus not optimized for embedding spaces. Correspondingly, the problem might be remedied by adding hypernyms to the labels, e.g. "hot dog (food)".\\

Training exclusively on text embeddings and a 65/15-COCO-split aligns the classification output to the CLIP embedding space at exactly 65 points.

This is because the CLIP model outputs exactly one text embedding for every distinct class label. 
Models additionally trained on image embeddings, such as ZSD-YOLO and ViLD, are aligned to the CLIP embedding space in as many points as there are different training images. This is because the CLIP model outputs a distinct image embedding for every image.
We hypothesize that the different image embeddings are clustered closely together and in close proximity to the corresponding text embeddings, while text embeddings of new \mbox{ImageNet} classes are spread further apart over the embedding space. 
This explains why we are able to more than double the performance gain of previous work even by adding only comparatively few text embeddings.

\section{Conclusion}
\label{sec:conclusion}
We present a simple method to cheaply improve zero-shot detection performance on unseen classes. Our experiments show that image labels can be used to better align the detector output to the embedding space. By using image labels in addition to the detection data set we are able to improve ZSD performance to a greater extent compared to previous work based on additional image embeddings. Adding ImageNet data benefits the training as the added data augments the training data set not only in quantity but also in quality, in form of new classes encountered during training.
Our method can also be used as a way of fine-tuning already trained models if they are trained to output embeddings.\\
Future work might develop the alignment to useful embedding spaces not by changing the data source, but by changing the method of alignment, e.g., the training loss. 
{\small
\bibliographystyle{ieee_fullname}
\bibliography{egbib}

\begin{thebibliography}{10}\itemsep=-1pt

\bibitem{zeroshot}
Ankan Bansal, Karan Sikka, Gaurav Sharma, Rama Chellappa, and Ajay Divakaran.
\newblock Zero-shot object detection.
\newblock In {\em Proceedings of the European Conference on Computer Vision
  (ECCV)}, pages 384--400, 2018.

\bibitem{yolov4}
Alexey Bochkovskiy, Chien-Yao Wang, and Hong-Yuan~Mark Liao.
\newblock Yolov4: Optimal speed and accuracy of object detection.
\newblock {\em arXiv preprint arXiv:2004.10934}, 2020.

\bibitem{DETR}
Nicolas Carion, Francisco Massa, Gabriel Synnaeve, Nicolas Usunier, Alexander
  Kirillov, and Sergey Zagoruyko.
\newblock End-to-end object detection with transformers.
\newblock In {\em European conference on computer vision}, pages 213--229.
  Springer, 2020.

\bibitem{R-FCN}
Jifeng Dai, Yi Li, Kaiming He, and Jian Sun.
\newblock R-fcn: Object detection via region-based fully convolutional
  networks.
\newblock {\em Advances in neural information processing systems}, 29, 2016.

\bibitem{devise}
Andrea Frome, Greg~S Corrado, Jon Shlens, Samy Bengio, Jeff Dean, Marc'Aurelio
  Ranzato, and Tomas Mikolov.
\newblock Devise: A deep visual-semantic embedding model.
\newblock {\em Advances in neural information processing systems}, 26, 2013.

\bibitem{OTA}
Zheng Ge, Songtao Liu, Zeming Li, Osamu Yoshie, and Jian Sun.
\newblock Ota: Optimal transport assignment for object detection.
\newblock In {\em Proceedings of the IEEE/CVF Conference on Computer Vision and
  Pattern Recognition}, pages 303--312, 2021.

\bibitem{yolox}
Zheng Ge, Songtao Liu, Feng Wang, Zeming Li, and Jian Sun.
\newblock Yolox: Exceeding yolo series in 2021.
\newblock {\em arXiv preprint arXiv:2107.08430}, 2021.

\bibitem{fastRCNN}
Ross Girshick.
\newblock Fast r-cnn.
\newblock In {\em Proceedings of the IEEE international conference on computer
  vision}, pages 1440--1448, 2015.

\bibitem{RCNN}
Ross Girshick, Jeff Donahue, Trevor Darrell, and Jitendra Malik.
\newblock Rich feature hierarchies for accurate object detection and semantic
  segmentation.
\newblock In {\em Proceedings of the IEEE conference on computer vision and
  pattern recognition}, pages 580--587, 2014.

\bibitem{vild-paper}
Xiuye Gu, Tsung-Yi Lin, Weicheng Kuo, and Yin Cui.
\newblock Open-vocabulary object detection via vision and language knowledge
  distillation.
\newblock {\em arXiv preprint arXiv:2104.13921}, 2021.

\bibitem{maskRCNN}
Kaiming He, Georgia Gkioxari, Piotr Doll{\'a}r, and Ross Girshick.
\newblock Mask r-cnn.
\newblock In {\em Proceedings of the IEEE international conference on computer
  vision}, pages 2961--2969, 2017.

\bibitem{SPP-Net}
Kaiming He, Xiangyu Zhang, Shaoqing Ren, and Jian Sun.
\newblock Spatial pyramid pooling in deep convolutional networks for visual
  recognition.
\newblock {\em IEEE transactions on pattern analysis and machine intelligence},
  37(9):1904--1916, 2015.

\bibitem{batchNormalization}
Sergey Ioffe and Christian Szegedy.
\newblock Batch normalization: Accelerating deep network training by reducing
  internal covariate shift.
\newblock In {\em International conference on machine learning}, pages
  448--456. PMLR, 2015.

\bibitem{ALIGN}
Chao Jia, Yinfei Yang, Ye Xia, Yi-Ting Chen, Zarana Parekh, Hieu Pham, Quoc Le,
  Yun-Hsuan Sung, Zhen Li, and Tom Duerig.
\newblock Scaling up visual and vision-language representation learning with
  noisy text supervision.
\newblock In {\em International Conference on Machine Learning}, pages
  4904--4916. PMLR, 2021.

\bibitem{FPN}
Tsung-Yi Lin, Piotr Doll{\'a}r, Ross Girshick, Kaiming He, Bharath Hariharan,
  and Serge Belongie.
\newblock Feature pyramid networks for object detection.
\newblock In {\em Proceedings of the IEEE conference on computer vision and
  pattern recognition}, pages 2117--2125, 2017.

\bibitem{RetinaNet}
Tsung-Yi Lin, Priya Goyal, Ross Girshick, Kaiming He, and Piotr Doll{\'a}r.
\newblock Focal loss for dense object detection.
\newblock In {\em Proceedings of the IEEE international conference on computer
  vision}, pages 2980--2988, 2017.

\bibitem{coco}
Tsung-Yi Lin, Michael Maire, Serge Belongie, James Hays, Pietro Perona, Deva
  Ramanan, Piotr Doll{\'a}r, and C~Lawrence Zitnick.
\newblock Microsoft coco: Common objects in context.
\newblock In {\em European conference on computer vision}, pages 740--755.
  Springer, 2014.

\bibitem{SSD}
Wei Liu, Dragomir Anguelov, Dumitru Erhan, Christian Szegedy, Scott Reed,
  Cheng-Yang Fu, and Alexander~C Berg.
\newblock Ssd: Single shot multibox detector.
\newblock In {\em European conference on computer vision}, pages 21--37.
  Springer, 2016.

\bibitem{SwinTransformer}
Ze Liu, Yutong Lin, Yue Cao, Han Hu, Yixuan Wei, Zheng Zhang, Stephen Lin, and
  Baining Guo.
\newblock Swin transformer: Hierarchical vision transformer using shifted
  windows.
\newblock In {\em Proceedings of the IEEE/CVF International Conference on
  Computer Vision}, pages 10012--10022, 2021.

\bibitem{wordnet}
George~A Miller.
\newblock Wordnet: a lexical database for english.
\newblock {\em Communications of the ACM}, 38(11):39--41, 1995.

\bibitem{conSE}
Mohammad Norouzi, Tomas Mikolov, Samy Bengio, Yoram Singer, Jonathon Shlens,
  Andrea Frome, Greg~S Corrado, and Jeffrey Dean.
\newblock Zero-shot learning by convex combination of semantic embeddings.
\newblock {\em arXiv preprint arXiv:1312.5650}, 2013.

\bibitem{gloVE}
Jeffrey Pennington, Richard Socher, and Christopher~D Manning.
\newblock Glove: Global vectors for word representation.
\newblock In {\em Proceedings of the 2014 conference on empirical methods in
  natural language processing (EMNLP)}, pages 1532--1543, 2014.

\bibitem{DetectoRS}
Siyuan Qiao, Liang-Chieh Chen, and Alan Yuille.
\newblock Detectors: Detecting objects with recursive feature pyramid and
  switchable atrous convolution.
\newblock In {\em Proceedings of the IEEE/CVF Conference on Computer Vision and
  Pattern Recognition}, pages 10213--10224, 2021.

\bibitem{clip-paper}
Alec Radford, Jong~Wook Kim, Chris Hallacy, Aditya Ramesh, Gabriel Goh,
  Sandhini Agarwal, Girish Sastry, Amanda Askell, Pamela Mishkin, Jack Clark,
  et~al.
\newblock Learning transferable visual models from natural language
  supervision.
\newblock In {\em International Conference on Machine Learning}, pages
  8748--8763. PMLR, 2021.

\bibitem{65split}
Shafin Rahman, Salman Khan, and Nick Barnes.
\newblock Improved visual-semantic alignment for zero-shot object detection.
\newblock {\em Proceedings of the AAAI Conference on Artificial Intelligence},
  34(07):11932--11939, Apr. 2020.

\bibitem{dlwl}
Vignesh Ramanathan, Rui Wang, and Dhruv Mahajan.
\newblock Dlwl: Improving detection for lowshot classes with weakly labelled
  data.
\newblock In {\em Proceedings of the IEEE/CVF conference on computer vision and
  pattern recognition}, pages 9342--9352, 2020.

\bibitem{yolo}
Joseph Redmon, Santosh Divvala, Ross Girshick, and Ali Farhadi.
\newblock You only look once: Unified, real-time object detection.
\newblock In {\em Proceedings of the IEEE conference on computer vision and
  pattern recognition}, pages 779--788, 2016.

\bibitem{yolo9000}
Joseph Redmon and Ali Farhadi.
\newblock Yolo9000: better, faster, stronger.
\newblock In {\em Proceedings of the IEEE conference on computer vision and
  pattern recognition}, pages 7263--7271, 2017.

\bibitem{yolov3}
Joseph Redmon and Ali Farhadi.
\newblock Yolov3: An incremental improvement.
\newblock {\em arXiv preprint arXiv:1804.02767}, 2018.

\bibitem{fasterRCNN}
Shaoqing Ren, Kaiming He, Ross Girshick, and Jian Sun.
\newblock Faster r-cnn: Towards real-time object detection with region proposal
  networks.
\newblock {\em Advances in neural information processing systems}, 28, 2015.

\bibitem{1000ImageNet}
Olga Russakovsky, Jia Deng, Hao Su, Jonathan Krause, Sanjeev Satheesh, Sean Ma,
  Zhiheng Huang, Andrej Karpathy, Aditya Khosla, Michael Bernstein,
  Alexander~C. Berg, and Li Fei-Fei.
\newblock {ImageNet Large Scale Visual Recognition Challenge}.
\newblock {\em International Journal of Computer Vision (IJCV)},
  115(3):211--252, 2015.

\bibitem{twostagevsonstage}
Petru Soviany and Radu~Tudor Ionescu.
\newblock Optimizing the trade-off between single-stage and two-stage deep
  object detectors using image difficulty prediction.
\newblock In {\em 2018 20th International Symposium on Symbolic and Numeric
  Algorithms for Scientific Computing (SYNASC)}, pages 209--214. IEEE, 2018.

\bibitem{EfficientDet}
Mingxing Tan, Ruoming Pang, and Quoc~V Le.
\newblock Efficientdet: Scalable and efficient object detection.
\newblock In {\em Proceedings of the IEEE/CVF conference on computer vision and
  pattern recognition}, pages 10781--10790, 2020.

\bibitem{transformers}
Ashish Vaswani, Noam Shazeer, Niki Parmar, Jakob Uszkoreit, Llion Jones,
  Aidan~N Gomez, {\L}ukasz Kaiser, and Illia Polosukhin.
\newblock Attention is all you need.
\newblock {\em Advances in neural information processing systems}, 30, 2017.

\bibitem{zsd-yolo}
Johnathan Xie and Shuai Zheng.
\newblock Zsd-yolo: Zero-shot yolo detection using vision-language
  knowledgedistillation.
\newblock {\em arXiv preprint arXiv:2109.12066}, 2021.

\bibitem{surveyObjectDetection}
Syed Sahil~Abbas Zaidi, Mohammad~Samar Ansari, Asra Aslam, Nadia Kanwal,
  Mamoona Asghar, and Brian Lee.
\newblock A survey of modern deep learning based object detection models.
\newblock {\em Digital Signal Processing}, page 103514, 2022.

\bibitem{ovr-cnn}
Alireza Zareian, Kevin~Dela Rosa, Derek~Hao Hu, and Shih-Fu Chang.
\newblock Open-vocabulary object detection using captions.
\newblock In {\em Proceedings of the IEEE/CVF Conference on Computer Vision and
  Pattern Recognition}, pages 14393--14402, 2021.

\bibitem{mixup}
Hongyi Zhang, Moustapha Cisse, Yann~N Dauphin, and David Lopez-Paz.
\newblock mixup: Beyond empirical risk minimization.
\newblock {\em arXiv preprint arXiv:1710.09412}, 2017.

\bibitem{contrastiveLoss}
Yuhao Zhang, Hang Jiang, Yasuhide Miura, Christopher~D Manning, and Curtis~P
  Langlotz.
\newblock Contrastive learning of medical visual representations from paired
  images and text.
\newblock {\em arXiv preprint arXiv:2010.00747}, 2020.

\bibitem{DeformableDETR}
Xizhou Zhu, Weijie Su, Lewei Lu, Bin Li, Xiaogang Wang, and Jifeng Dai.
\newblock Deformable detr: Deformable transformers for end-to-end object
  detection.
\newblock {\em arXiv preprint arXiv:2010.04159}, 2020.

\end{thebibliography}
}

\appendix
\section{class filtering and promp engineering}
\label{appendix}
From the 1000 class ImageNet data set \cite{1000ImageNet}, we exclude the following classes: airplane wing, airliner, military aircraft, high-speed train, 
    parking meter, tabby cat, tiger cat, Persian cat, Siamese cat, Egyptian Mau, 
    brown bear, American black bear, polar bear, sloth bear, hot dog, toilet seat, 
    computer mouse, toaster, hair dryer.\\
Classes are excluded from training if their usage would lead to an unfair advantage on the unseen COCO classes during testing. This leads not only to the exclusion of ImageNet classes that are identical to ("parking meter") or synonyms of ("airliner") unseen COCO classes. For example, we also exclude the ImageNet class "Persian cat" because it is a hyponym of the unseen COCO class "cat", and we exclude the ImageNet class "airplane wing" because it is a meronym of airplane, an unseen COCO class.\\
The seven templates\footref{clip-embeddings-git} that were used to precompute the CLIP embeddings are:
\begin{verbatim}
"itap of a {class}."
"a bad photo of the {class}."
"a origami {class}."
"a photo of the large {class}."
"a {class} in a video game."
"art of the {class}."
"a photo of the small {class}."
\end{verbatim}

\section{smaller model size}\label{sec:additional_experiments}
\begin{table}
  \centering
  \begin{tabular}{lclc}
    \toprule
    Model & mAP &\\
    \midrule
    Ours size s (w/o image labels) & 7.4 & \\
    Ours size s & 9.5 & \textbf{+2.1} \\
    \bottomrule
  \end{tabular}
  \caption{mAP (in \%)  on the unseen classes of the 65/15 COCO split after training our model on 10\% of COCO and jointly on 10\% of COCO and 10\% of ImageNet. The benefit of adding ImageNet (2.1 points mAP) is smaller than with a larger model of the YOLOX family.}
  \label{tab:imagenet-gain}
\end{table}

We train a modified version of a smaller YOLOX model (YOLOX-s) with the Mosaic data augmentation.

Table \ref{tab:imagenet-gain} shows the mAP after training on 10\% of COCO data. Adding 10\% of ImageNet in this case increases the mAP by 2.1 percentage points. This is less than the increase of 3.3 percentage points when training the larger model on the full datasets.

\begin{figure}
  \includegraphics[width=\linewidth]{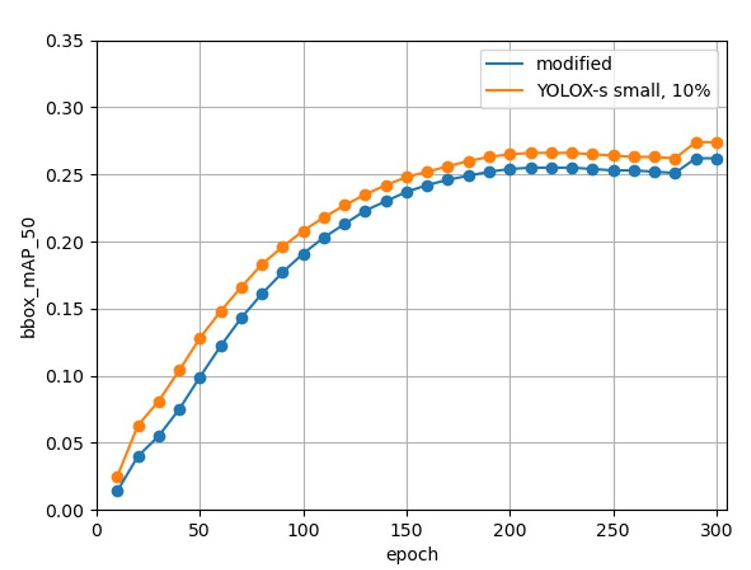}

  \caption{mAP evaluated on seen classes of two YOLOX-s models trained on the same data (10\% of COCO) with the same augmentations (Mixup and Mosaic), one with a modified classification head and one with the original YOLOX implementation. The model with the original YOLOX implementation achieves 27.4 mAP and the modified model achieves 26.2 mAP after 300 epochs. Apart from this slight drop in mAP the overall learning progress is the same.}
  \label{fig:models-s}
\end{figure}

In \ref{fig:models-s} we compare the results to those of an unmodified YOLOX model of the same size trained with the same data augmentation. The learning of the seen classes follow the same overall trajectory, with a slight drop in performance that is to be expected when switching to embeddings and thus fixing the classifier. Our changes in the classification head do not have major influence on the mAP.
\end{document}